\documentclass[a4paper]{article}

\usepackage{iwslt18,amssymb,amsmath,epsfig}

\usepackage{times}
\usepackage{latexsym}
\usepackage{url}

\usepackage[english]{babel}
\usepackage[utf8x]{inputenc}

\usepackage{hyphenat}
\usepackage{amsmath}
\usepackage{graphicx}
\usepackage{float}
\usepackage[justification=justified]{caption}
\usepackage{enumitem} 
\usepackage{dirtytalk}
\usepackage{array}
\usepackage{multirow}

\usepackage{siunitx}
\usepackage{color}

\newcommand{\citet}{\cite}
\newcommand{\citep}{\cite}

\title{The ADAPT System Description for the IWSLT 2018 Basque to English Translation Task }

\twoauthors{Alberto Poncelas, Andy Way}
        {ADAPT Centre, School of Computing, \\
        Dublin City University, Dublin, Ireland \\
        \tt \{firstname.lastname\}@adaptcentre.ie}
        {Kepa Sarasola}
        {Ixa Group (UPV/EHU), Faculty of Informatics \\
         University of the Basque Country \\
         \tt kepa.sarasola@ehu.eus}

\begin{document}

\maketitle

\begin{abstract}

In this paper we present the ADAPT system built for the Basque to English Low Resource MT Evaluation Campaign. Basque is a low-resourced, morphologically-rich language. This poses a challenge for Neural Machine Translation models which usually achieve better performance when trained with large sets of data.

Accordingly, we used synthetic data to improve the translation quality produced by a model built using only authentic data. Our proposal uses back-translated data to: (a) create new sentences, so the system can be trained with more data; and (b) translate sentences that are close to the test set, so the model can be fine-tuned to the document to be translated.

\end{abstract}

\section{Introduction}

We participated in the Basque to English Low Resource MT Evaluation Campaign as part of the International Workshop on Spoken Language Translation (IWSLT) 2018. In this task, we aimed to build an MT model to translate subtitles of TED (Technology, Entertainment, Design) talks from Basque into English.

Basque (or Euskera), which is mainly spoken in the Basque Country in Northern Spain and Southern France, is considered an isolated language. Linguistically, it is an agglutinative language, and morphologically more complex than English. Furthermore, Basque is a low resource language. Due to these characteristics, creating an MT  system  that deals with Basque is a challenging task.

As the MT Evaluation Campaign consists of translating subtitles from TED talks, we built our MT engines mainly from available subtitles. TED Talks\footnote{\url{https://www.ted.com}} is an event where experts in different fields, such as education, business, science, etc. give a talk of up to 18 minutes to disseminate their ideas.

The use of subtitles as training data is potentially problematic as they may not be literal translation, causing the original and translated sentences not to be truly parallel. This is because subtitles are subjected to a great deal of adaptation. Localization strategies (adapting the text to suit consumers of a particular locale or culture), combined with the requirement to meet time constraints (where sentences in the source and target languages which have different length are supposed to appear on the screen within the same time frame), results in sentences which are comparable but not necessarily parallel \citep{fishel2012subtitles}.

Although the adaptation does not hinder human comprehension of the intended message, when these sentences are used as training data for an MT model, the translation inaccuracies become obstacles for the system to correctly learn to translate.

The system presented in this paper aims to overcome the aforementioned problems. First, the creation of synthetic data has two purposes: (i) it provides a new set of parallel sentences that mitigates the problem of Basque being a low resourced language; and (ii), artificially-created sentences tend to be more literal than usual translated subtitles. Therefore the former may constitute better training data for an MT model than the latter. Secondly, as TED Talks topics cover a wide variety of domains, we use data selection techniques to adapt an MT model to a particular test set.

The remainder of the paper is structured as follows. In Section \ref{sec:related_work}, we describe related work regarding MT models that include Basque as source or target language. We also describe previous work on the use of synthetic data and data selection algorithms that are related to the systems described below. Section \ref{sec:system_description} describes the two steps (hybrid corpus creation and model adaptation) performed for building the MT system. In Section \ref{sec:experimental_results} we present an estimation of the performance of the models created. Finally, an overview of the system is described in Section \ref{sec:conclusions}.

\section{Related Work}
\label{sec:related_work}

The system described in this paper is based on two main techniques: (a) incorporating synthetic sentences as training data (Section \ref{sec:addition_synth_data}), and (b) adapting the model to the test set (Section \ref{sec:adapting_model}).

\subsection{Basque Machine Translation}

Most of the work on MT involving Basque is based on the Basque-Spanish pair. We can find multiple MT approaches including Rule Based MT (RBMT) \cite{mayor2011matxin}, or data-driven approaches \citep{labaka2007comparing} such as Example-based MT \citep{stroppa2006example} or hybrid (Statistical MT and RBMT) \citep{labaka2014hybrid} systems.

Dealing with low-resource languages is a problem for NMT approaches as they require large amounts of data in order to generate good translations. For some language pairs, SMT models can outperform NMT models when trained in limited amount of data \citep{dowling2018smt}. In the work of Unanue et al. (2018) \citep{unanue2018english} they perform a comparison of Basque\hyp English SMT and NMT models. Their finding reveals that SMT models trained with \textit{PaCo2-EuEn} corpus in the Basque-to-English direction perform better than NMT models. In the reverse direction, however, NMT models can perform better when pre-trained embeddings (which have been trained using additional sentences from Basque Wikipedia) are given to the model.

Regarding Basque-Spanish NMT models, the most recent work is presented by Etchegoyen et al. (2018) \citep{EtchegoyenEAMT:18} where they explored different methods of splitting words into morphemes to improve the translation.

\subsection{Addition of Back-translated Sentences}
\label{sec:addition_synth_data}

As Basque is a low-resource language, the amount of available parallel data is very limited. A technique to increase the number of sentences is to artificially create sentences. Sennrich et al. (2016) \cite{sennrich2016improving}, showed that NMT models could be boosted by adding backtranslated data. 

Backtranslation in this paper designates the process of translating monolingual sentences in the target language into the source-side language. By doing this, a synthetic parallel corpus is created. Adding this corpus as training data can improve the performance of the model. In fact, models built using solely back-translated data can even achieve comparable performance to those trained with authentic or hybrid data \citep{poncelas2018investigating}.

\subsection{Adaptation of the MT Model to the Test Set}
\label{sec:adapting_model}

There are several techniques for adapting a model to a particular domain \citep{chu2018survey}, such as selecting relevant data (\textit{data-centric} approaches), or modifying the model (\textit{model-centric} approaches). 

In the case where the test set is available, it is possible to adapt the model so it performs better in the given test. In our work, we used a combination of data-centric and model-centric approaches. First, we selected data that are relevant for the test set, and then we used fine-tuning to bias the model towards the test set.

Fine-tuning \cite{luong2015stanford,freitag2016fast}, consists of using a pre-built NMT model (trained on general domain data), and training the last epochs on smaller amounts of in-domain data. An alternative to this technique is \textit{gradual fine tuning} \cite{van2017dynamic}, which involves reducing the training data as the training proceeds.


While these fine-tuning techniques aim to adapt the NMT models towards a particular domain, Li et al. (2018) \citep{li2018one} proposed to use fine-tuning to adapt the model to the test set, which is closer to our approach. The main difference is that while in their work the model is adapted sentence-wise (one model for each sentence), in ours, it is adapted document-wise (one model for the document).

In order to select sentences that are closer to the test set we used Feature Decay Algorithms (FDA) \citep{biccici2011instance,bicici2015parfda,biccici2015optimizing}. This technique has been successfully applied in both SMT \citep{biccici2013feature,poncelasextending,poncelas2017applying} and NMT \citep{poncelas2018feature}. 

FDA is a data selection method that not only aims to select sentences that are close to a seed (generally the test set), but also to promote the variability of the training data selected.

In order to achieve that, FDA scores each sentence $s$ in the parallel data, and the sentence with the highest score is added to a list of selected sentences $L$. The score of the sentence is based on how similar it is to the seed (counting the {\em n}-grams shared with the seed), and how different it is to previously selected sentences (penalizing {\em n}-grams already contained in $L$), which increases the variability.

Using default values of the parameters, the score of a sentence is computed as in Equation \eqref{eq:fda_sentencescore_general}:

\begin{equation}\label{eq:fda_sentencescore_general}
score(s|L)=\frac{\sum_{ngr \in s} 0.5^{C_L(ngr)}}{length(s)}
\end{equation}

\noindent where $C_L(ngr)$ is the count of the {\em n}-gram $ngr$ in the pool of selected sentences $L$. The more occurrences of $ngr$ there are in $L$ the more penalized $ngr$ is. The factor $0.5^{C_L(ngr)}$ in Equation \eqref{eq:fda_sentencescore_general} causes the {\em n}-gram to contribute less to the total score of the sentence.

\section{System Description}
\label{sec:system_description}

The system built consists of two steps. First, (Section \ref{sec:addition_synthetic_data}) we created a basic model using authentic and synthetic data. In the second step (Section \ref{sec:adapt_base_model}), the model was fine-tuned to be adapted to the test set.

\subsection{Basque-English Data}

The Basque-English parallel data used in this work were obtained by combining the OpenSubtitles2016 (173K sentences), OpenSubtitles2018 \citep{Tiedemann:RANLP5} (805K sentences) and the \textit{PaCo2-EuEn} corpus\footnote{\url{komunitatea.elhuyar.org/ig/files/2016/01/PaCo_EuEn_corpus.tgz}} \citep{san2012paco2} (130K sentences) provided in the IT domain MT Shared Task of WMT 2016 \citep{bojar2016findings}. We randomly sampled 5000 sentences as our dev set and the rest (1M sentences) as the training set.

In order to build the NMT models we used OpenNMT-py, which is the Pytorch port of OpenNMT \citep{opennmt}. All the NMT models we built were configured with the same settings (the only difference is the training data used to build them). The value parameters were the default ones of OpenNMT-py (i.e. 2-layer LSTM with 500 hidden  units, vocabulary size of 50000 words for each language). All the models were executed for 13 epochs, and we also used Byte Pair Encoding (BPE) \citep{sennrich2016neural} with 30000 merge operations, following the work of Etchegoyen et al. (2018) \citep{EtchegoyenEAMT:18}.

\subsection{Addition of Synthetic Data}
\label{sec:addition_synthetic_data}

\begin{figure}[hbt]
\includegraphics[width=8cm, height=6cm]{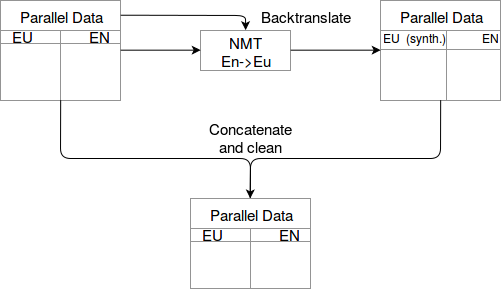}
\centering
\caption{ Creation of hybrid parallel corpus.
 \label{fig:hybrid_data_creation}
}
\end{figure}

The first step in the construction of a baseline system is to extend the parallel corpus. In Figure \ref{fig:hybrid_data_creation} we present a diagram of how we built the corpus. Using an initial corpus of parallel Basque-English sentences we built an NMT model capable of translating sentences from English into Basque. Then, the English side of that parallel corpus was translated into Basque using the English to Basque NMT model. 


Intuitively, translating the same sentences that were fed as training data should not be useful as it is likely to produce very similar sentences. However, the sentences produced by the model tend to be more literal translations, thereby avoiding the problems previously mentioned.

In Table \ref{table:examples_synth sentences} we show some examples of how synthetic data present Basque sentences that are closer to literal translation than a human-produced sentences. For example, in the first row, the translation for the English sentence \say{do I need to be there?} is \say{joan behar dut?}, which literally means \say{do I have to go?}. The artificially-created sentence is a more precise translation, as it uses the verb \say{be} (\say{egon}) instead of the verb \say{go} (\say{joan}). In certain contexts, the use of one or another sentence does not affect the general understanding. However, using the wrong translation as training data for a model can hurt performance.

A similar effect is observed in the second row of Table \ref{table:examples_synth sentences} for the sentence \say{keep her steady, now.}. The Basque translation of this sentence is \say{ez dadila mugitu.} which uses the verb \say{mugitu} (\say{to move}), so it could be translated as \say{it shall not move} or \say{do not let it move}. In contrast, the MT model produced the sentence, \say{eutsi gogor.}, which used the verb \say{hold} (\say{eutsi}). Both translations are appropriate, but they belong to different contexts.

Finally, we see in the third row the English sentence \say{a suicide?}. The corresponding sentence in Basque is \say{nor zen?} (\say{who was?}). In any other context, the two sentences have completely different meanings. The synthetic sentence by contrast is a literal translation.



\begin{table}[!htbp]
\centering
\begin{center}
\begin{tabular}{ |p{0.1cm}|p{2.1cm}|p{2.2cm}|p{2.1cm}|}
\hline
& Authentic Basque	&   Synthetic Basque	&	English\\
\hline
1&joan behar dut?	&	hor egon behar dut?	&	do I need to be there?\\
2&ez dadila mugitu.	&	eutsi gogor.	&	keep her steady, now.\\
3&nor zen?	&	suizidioa?	&	a suicide?\\
\hline
\end{tabular}
\caption{Examples of sentences in Basque (authentic), Basque (synthetic) and English translation. }
\label{table:examples_synth sentences}
\end{center}
\end{table}


Following backtranslation we obtained two parallel sets, with authentic and synthetic sentences. Next, we concatenated them as a single corpus. Note that, by doing so, the target-language sentences are duplicated.

Finally, we removed those sentences in which the length of the source and target sides differed substantially. In our work we kept a sentence pair $(s,t)$ if $0.5 < \frac{len(s)}{len(t)}>1.5$, in order to remove the 10\% outliers. In total 255K sentences were removed (137K sentences 118K sentences from authentic  and synthetic sets, respectively). The hybrid corpus contained, therefore, 1.93M sentence pairs.

We applied these criteria to both corpus of authentic, and synthetic sentences, so the potentially unaligned sentences are ignored and bad translated sentences are not considered, respectively.


\subsection{Adaptation to the Test Set}
\label{sec:adapt_base_model}

\begin{figure}[hbt]
\includegraphics[width=8cm, height=9cm]{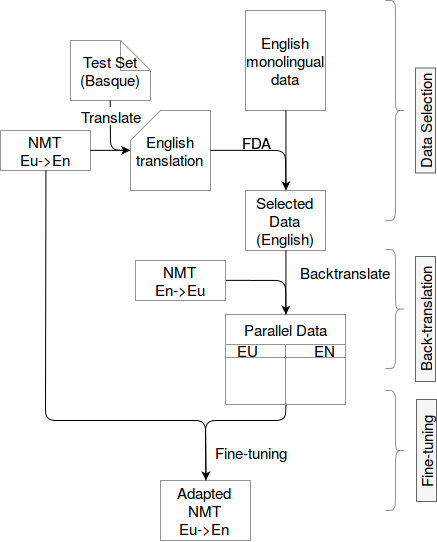}
\centering
\caption{ Fine tuning with synthetic data
 \label{fig:Synth_FineTuning}
}
\end{figure}

The second step of building the model is to adapt it to a particular test set. The work of Poncelas et al. (2018) \cite{poncelas2018feature} showed that when the test set is available during training time it is possible to fine-tune a model to improve the translation of that particular test set.

In Figure \ref{fig:Synth_FineTuning} we show how we fine-tuned our NMT model, which requires three phases as follows:
\begin{enumerate}
    \item Data Selection: In this phase we aimed to retrieve English sentences that were close to the test set. As the test set was in Basque, we first created an approximated translation using the NMT model built as explained in Section \ref{sec:addition_synthetic_data}. This translation can be used as the seed for FDA and extract a set of sentences, from a monolingual English corpus, that were close to the pre-translated set and hence, to the test set. In this work we extracted 50,000 sentences from English training data provided in the WMT 2015 Translation Task \citep{bojar-EtAl:2015:WMT}.
    \item Back-translation: The subset of selected English sentences were back-translated (we reused the same English\hyp to\hyp Basque model built as explained in Section \ref{sec:addition_synth_data} to create backtranslated data) in order to build a parallel corpus.
    \item Fine-tuning: The synthetic parallel corpus was used to fine-tune the MT model for one epoch. In this way, the model was tailored to the test set.
\end{enumerate}

\section{Experimental Results}
\label{sec:experimental_results}

In order to estimate the performance of the final and intermediate models described through Section \ref{sec:system_description} we evaluated them using the development set (containing 1K sentences extracted from subtitles of TED talks) provided by the organizers of the IWSLT Evaluation Campaign.

The models evaluated are: (a) the model built with only authentic data (\textit{base model}); (b) the model built with the combination of authentic and synthetic data (\textit{hybrid model}); and (c), the \textit{hybrid model} adapted to the test set using FDA-retrieved data (\textit{FDA model}).

\begin{table}[!htbp]
\centering
\begin{center}
\begin{tabular}{ |p{1.3cm}|p{1.6cm}|p{1.6cm}|p{1.6cm}|}
\hline
	&	\textit{base model}	&	\textit{hybrid model} & \textit{FDA model}	\\
\hline	
BLEU	&	0.1315	&	\bf0.1426*	&	\bf0.1450*	\\
NIST	&	4.459	&	\bf4.683	&	\bf4.733	\\
TER 	&	0.8508	&	0.8576	&	0.8666	\\
METEOR	&	0.1429	&	\bf0.1501*	&	\bf0.1528**	\\
CHRF3	&	34.05	&	\bf35.92	&	\bf36.24	\\
CHRF1	&	37.40	&	\bf38.67	&	\bf38.81	\\
\hline	
\end{tabular}
\caption{ 
Evaluation of the model built only with authentic data and using both authentic and synthetic data.
 \label{table:models_auth_synth_fda}}
\end{center}
\end{table}


We used several evaluation metrics to compare the outputs of the three models to a human-translated reference. In Table \ref{table:models_auth_synth_fda} we can see the evaluation scores for each model. The metrics we present are BLEU \citep{papineni2002bleu}, NIST \citep{doddington2002automatic}, 
TER \citep{snover2006study}, METEOR \citep{banerjee2005meteor} 
and CHRF3 \citep{popovic2015chrf}.


We also marked in bold the scores that outperform those of the \textit{base model} (first column of Table \ref{table:models_auth_synth_fda}) and marked with an asterisk the scores (among BLEU, TER and METEOR) that are statistically significant at level p=0.01. This was computed with multeval \citep{clark2011better} using Bootstrap Resampling \citep{koehn04}. The two asterisks in column \textit{FDA model} (METEOR row) indicate that it is statistically significant at p=0.01 when compared not only to the \textit{base model} but also to the \textit{hybrid model}.


As mentioned in Section \ref{sec:addition_synthetic_data}, the addition of synthetic data (even if it consists of a backtranslation of the same data used for training the model) is helpful. This is verified with the results in column \textit{hybrid model} in Table \ref{table:models_auth_synth_fda}. As we can see, most of the \textit{hybrid model} scores of the model are better than the model built with authentic data only (\textit{base model} column) and according to two of the scores, the improvements are statistically significant at p=0.01. In fact, a model built using only synthetic data (Table \ref{table:models_synth}) can achieve improvements over the \textit{base model}, according to METEOR and CHRF3 evaluation metric. 


\begin{table}[!htbp]
\centering
\begin{center}
\begin{tabular}{ |p{1.3cm}|p{1.6cm}|}
\hline
	&	\textit{synth. model} \\
\hline	
BLEU	&	0.1224	\\
NIST	&	4.074	\\
TER 	&	0.9769	\\
METEOR	&	\bf0.1481	\\
CHRF3	&	\bf36.22	\\
CHRF1	&	36.40	\\
\hline	
\end{tabular}
\caption{ 
Evaluation of the model built with synthetic data only.
 \label{table:models_synth}}
\end{center}
\end{table}

Finally, fine-tuning the \textit{hybrid model} using sentences that are close to the test set is also beneficial. As we can see in the column \textit{FDA model} (in Table \ref{table:models_auth_synth_fda}), most of the scores (except TER) are better than those of the \textit{base model} or even the \textit{hybrid model}, and according to METEOR metric the improvement is statistically significant at p=0.01.


\section{Conclusion}
\label{sec:conclusions}

In this paper we have described the ADAPT system presented for the Low Resource MT Evaluation Campaign of IWSLT 2018. The system translates from Basque into English.

Basque is a morphologically rich language, which causes the task of building an MT model to be more difficult than languages such as Spanish or German. Furthermore, the available parallel Basque-English data are scarce.

Due to the limited resources of texts in Basque, we generated synthetic data that successfully boosted the performance of the MT model trained solely with authentic sentences.

Additionally, we have used a supplementary monolingual English corpus so we could retrieve sentences close to the test set and further improve our model.

\section{Acknowledgments}
The research leading to these results was carried out as part of the TADEEP project (Spanish Ministry of Economy and Competitiveness TIN2015-70214-P, with FEDER funding).
This work has been supported by the ADAPT Centre for Digital Content Technology which is funded under the SFI Research Centres Programme (Grant 13/RC/2106) and is co-funded under the European Regional Development Fund.

\bibliographystyle{IEEEtran}
\bibliography{bibl}

\end{document}